\documentclass[conference]{IEEEtran}

\IEEEoverridecommandlockouts
\usepackage[utf8]{inputenc} 

\usepackage{cite}
\usepackage{amsmath,amssymb,amsfonts}
\usepackage{algorithmic}
\usepackage{graphicx}
\usepackage{textcomp}
\usepackage{caption}
\usepackage{subcaption}
\usepackage{float}
\usepackage{siunitx}
\usepackage{algorithm}
\usepackage{balance}

\begin{document}

\title{Prototyping of a multirotor UAV for precision landing under rotor failures
\thanks{This work was partially supported by the following projects: NVIDIA Applied Research Program Award 2021, UBACyT-2018 20020170100421BA  Universidad de Buenos Aires,  PICT-2019-2371 and PICT-2019-0373  Agencia Nacional de Investigaciones Cient\'ificas y Tecnol\'ogicas.}
}

\author{\IEEEauthorblockN{Alvaro J. Gaona}
\IEEEauthorblockA{\textit{LINAR} \\
\textit{Universidad de San Andrés}\\
San Fernando, Argentina\\
agaona@udesa.edu.ar}
\and
\IEEEauthorblockN{Claudio D. Pose}
\IEEEauthorblockA{\textit{Facultad de Ingeniería (FIUBA)} \\
\textit{Universidad de Buenos Aires and} \\
\textit{CONICET, Buenos Aires, Argentina} \\
cldpose@fi.uba.ar}
\and
\IEEEauthorblockN{Juan I. Giribet}
\IEEEauthorblockA{\textit{CONICET and LINAR} \\
\textit{Universidad de San Andrés}\\
San Fernando, Argentina\\
jgiribet@conicet.gov.ar}
\and
\IEEEauthorblockN{Roberto Bunge}
\IEEEauthorblockA{\textit{LINAR} \\
\textit{Universidad de San Andrés}\\
San Fernando, Argentina\\
rbunge@udesa.edu.ar}
}

\maketitle

\begin{abstract}
This work presents a prototype of a multirotor aerial vehicle capable of precision landing, even under the effects of rotor failures. The manuscript presents the fault-tolerant techniques and mechanical designs to achieve a fault-tolerant multirotor, and a vision-based navigation system required to achieve a precision landing. Preliminary experimental results will be shown, to validate on one hand the fault-tolerant control vehicle and, on the other hand, the autonomous landing algorithm. Also, a prototype of the fault-tolerant UAV is presented, capable of precise autonomous landing, which will be used in future experiments.

\end{abstract}

\begin{IEEEkeywords}
UAV, autonomous landing, fault tolerance, vision-based navigation.
\end{IEEEkeywords}

\section{Introduction}\label{sec:intro}

Recent decades have seen an exponential growth in the development and use of autonomous aerial vehicles, which are capable of transporting passengers and cargo, performing remote monitoring or dangerous tasks with a significant reduction in operating costs due to the reduced need for human operators, both on-board and on the ground. These vehicles have applications in domains as diverse as passenger transport and logistics, infrastructure monitoring, application to agriculture, early response to natural disasters and emergencies, communications relay, and provision of internet services, among many others. Globally, autonomous aerial vehicles represent a market of approximately USD 4 billion and it is projected that by 2030 will reach USD 24 billion \cite{AAM1}.

In the last decade, due to the progress of smaller, more powerful, and cheaper sensors and computers, and the convergence of distributed electric propulsion and storage technologies, in particular, multirotor type vehicles have become one of the better alternatives due to their maneuverability, being able to operate in small spaces and without requiring large dedicated infrastructures, thus enabling, among other things, operation in urban environments. The market potential for multirotor has driven the creation of a phenomenal number of technology-based companies dedicated to the development of these technologies, reaching a total of 10 billion in investment in the last 5 years, of which almost 5 billion correspond to the last year only \cite{AAMFund1}.

The autonomous operation of this type of aircraft requires addressing a set of problems, in particular, the landing phase is undoubtedly the one that represents the greatest risk, due to the proximity to the ground and the different infrastructures, and the possibility of imminent collision. As the operating space shrinks and operations take place in urban environments, the risk increases considerably. This is not only the case during a nominal flight, where all systems on-board and on the ground are operating within nominal parameters and the challenge is to land in a confined space without collision, but it is significantly aggravated in emergency cases, where the vehicle must execute a forced emergency landing in conditions of reduced maneuverability and uncontrolled spaces with the possibility of damage to third parties. This remains to this day a critical point, if not the most important, in the certification of autonomous aerial vehicles and is a central obstacle when it comes to bringing this type of product to the market \cite{NASAAAM}.

\begin{figure}[t]
    \centering
    \includegraphics[width = \columnwidth]{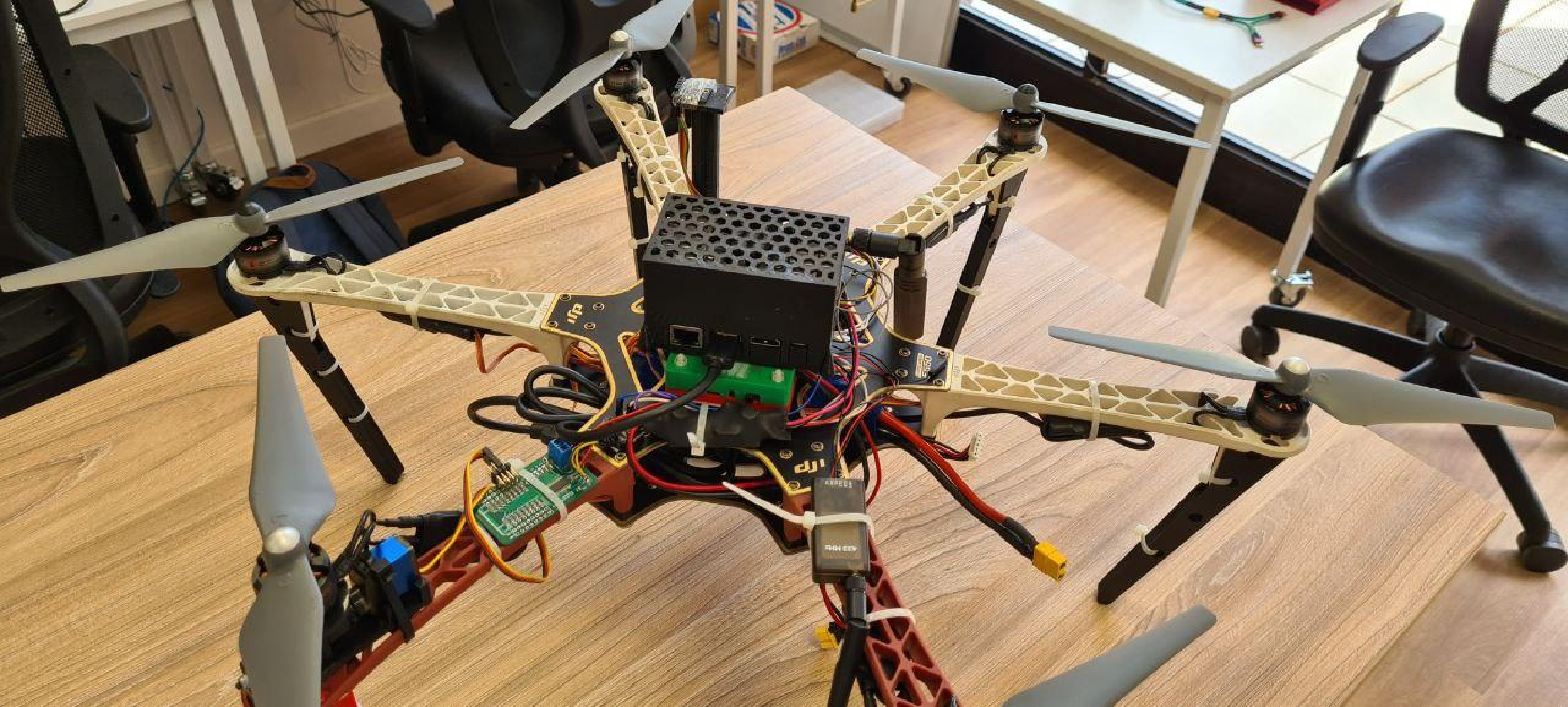}
    \caption{Hexarotor fault-tolerant vehicle. On the bottom-left arm, a servo allows to tilt the re-configurable motor. Two computers can be seen: inside a black case, the NVIDIA Jetson TX2, and inside a green case, the flight controller.}
    \label{fig:hexarotor_platform}
\end{figure}

In \cite{giribet2016analysis}, the capability of compensating for a rotor failure without losing the ability to exert torques in all directions, and therefore keeping full attitude control in case of failure, is studied. For this, at least six rotors are needed, and the proposed solution was to tilt the rotors inwards with respect to the vertical axis of the vehicle.
Experimental results for the proposed solution can be found in \cite{giribet2018experimental}; while the system proved to work correctly, there was a direction that, when exerted torque in, performed noticeably worse with respect to the rest. 
In \cite{Michieletto2017}, a detailed analysis is made for the optimal orientations of the rotors in a hexacopter, to achieve full tolerant attitude control. Still, the maximum torque achievable in some directions may be too small, with the consequent degradation of vehicle maneuverability. 

To overcome this limitation, in \cite{Pose2020TMech} several hexacopter structures and their fault-tolerant capabilities are analyzed. It is shown that a possible solution is to convert these structures into re-configurable ones, to significantly improve the maneuverability in case of a total failure in one rotor. 
By adding a mechanism to tilt one or two rotors sideways in case of a failure, a standard hexacopter vehicle can be converted into a fault-tolerant one. Furthermore, it was shown that it is enough to install this tilting mechanism in only two rotors, reducing the necessity of mechanical parts, improving the reliability of the vehicle.

We are currently working on the validation of this fault-tolerant hexacopter vehicle to achieve a precise autonomous landing. For this purpose, we developed the prototype presented in this work. This prototype consists of a standard hexacopter vehicle where a servomotor is installed. The servomotor is not used to control the vehicle, instead only tilts the rotor when a failure is detected. 
In order to achieve a precise landing maneuver, a visual navigation algorithm based on fiducial markers is used.

This work is organized as follows. First, the fault-tolerant hexacopter is presented and experimental results of this vehicle under failure conditions are presented. Next, the autonomous landing algorithm is presented. This algorithm is evaluated in a multirotor vehicle in nominal conditions, to test the performance. In the last section, the prototype, that will be used in the future to carry out the experimental validation for a precise autonomous landing, is presented.

\section{fault-tolerant vehicle design}\label{sec:ftcs}

Generally, unmanned aerial vehicles (UAVs) of the multirotor type are not fault-tolerant, in the sense of maintaining full attitude and altitude control after the occurrence of a failure in one of its actuators. It has been proved that, to achieve this, a minimum of six rotors is needed \cite{giribet2016analysis}, which draws attention to the particular case of hexarotor vehicles. It is still possible that vehicles with more rotors can achieve fault tolerance, considering solutions as simple as actuator redundancy, but those type of vehicles tend to be bigger in size and sometimes impractical. On the other hand, hexarotor vehicles provide a good balance of size and payload capacity, which is enough for the current most common applications.

Still, standard hexarotor vehicles, those with a structure design such that the motors are positioned in the vertices of a regular hexagon and pointing upwards, are not fault-tolerant. However, many solutions in hexarotor design have been proposed to achieve fault tolerance with six rotors. Particularly, it was shown in \cite{Pose2020TMech}, that reconfigurable tilting-rotor hexarotors can provide a very good degree of maneuverability in case of rotor failures. These solutions work on the following principle. If the vehicle presents no failures, the structure of the hexarotor is the same as the standard one, as it provides the best maneuverability with no energy waste. In case of motor failure, one of two rotors changes its orientation to a pre-computed optimal one; which of the two motors is reconfigured, and at which orientation, depends on the motor that presented the failure. Each one of the two reconfigurable motors is able to compensate the failure of three of the other motors.

For the prototype discussed here, a vehicle with such a structure was used, shown in Fig. \ref{fig:hexarotor_platform}. Here, only one of the motors was converted to reconfigurable, which allows to compensate three possible motor failures. As the experimental tests will be conducted making always fail the same motor, one reconfigurable motor is enough, since the results can be easily generalized to a failure in any rotor. 

\begin{figure}[t]
    \centering
    \includegraphics[width = \columnwidth]{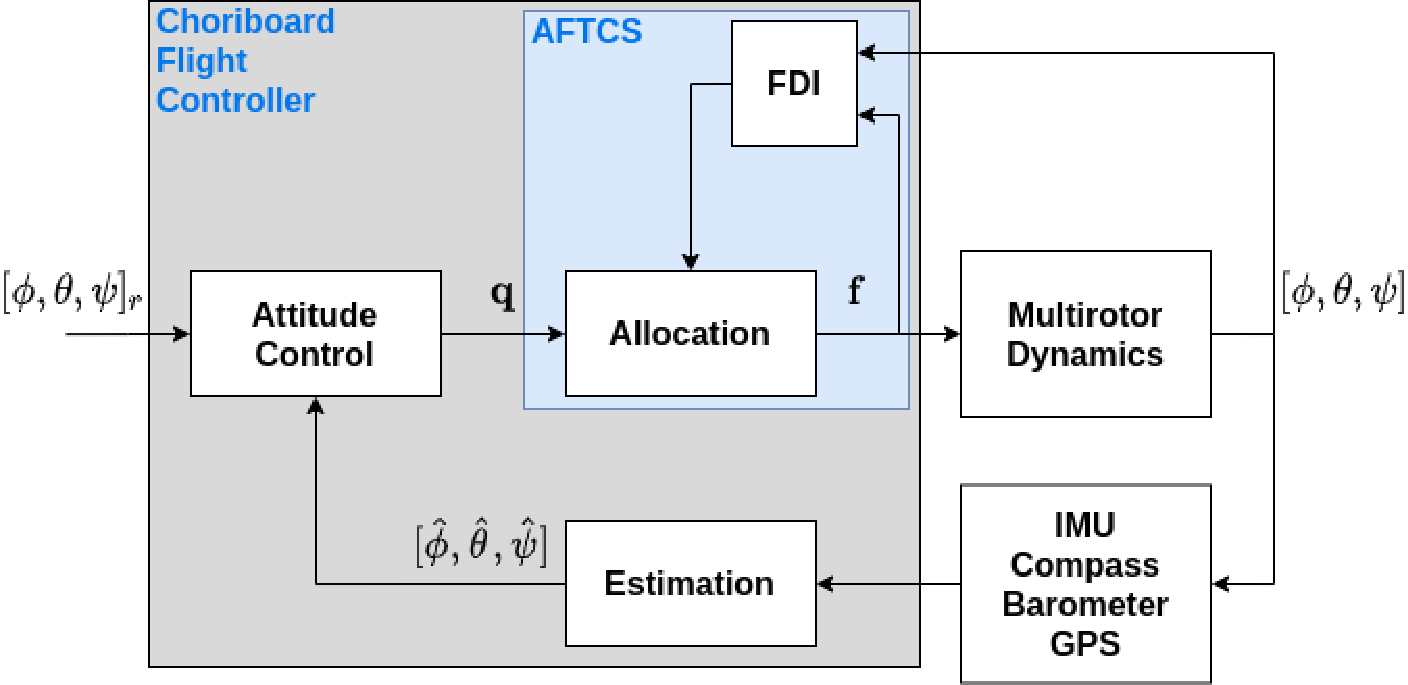}
    \caption{Architecture of the Fault Detection and Identification (FDI) module.}
    \label{fig:diag_FDI_v2}
\end{figure}

\begin{figure}[t]
    \centering
    \includegraphics[width = \columnwidth]{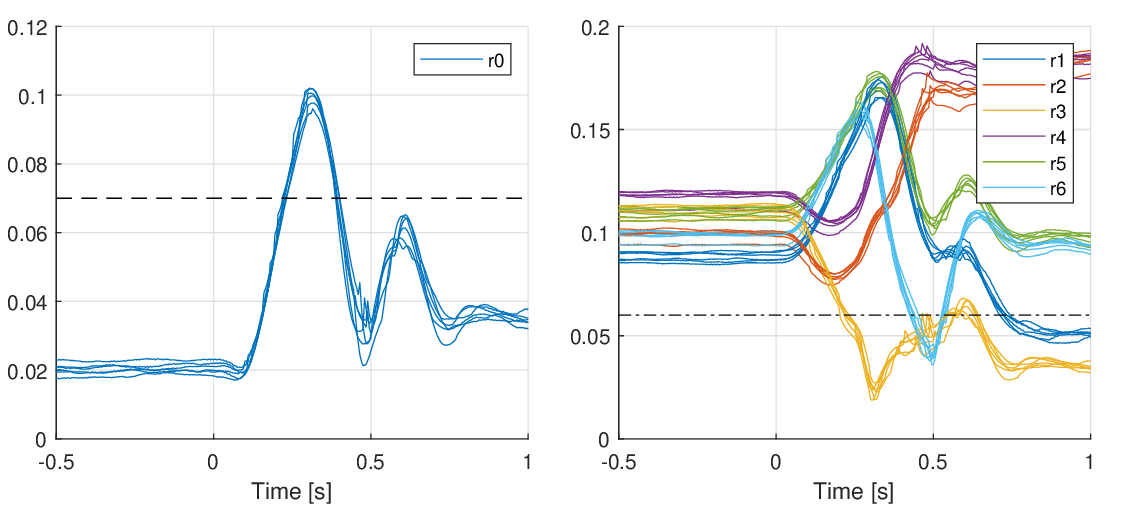}
    \caption{Detection (left) and identification (right) residues for several flights in which a failure is injected at $t=\SI{0}{\second}$ in motor 3.}
    \label{fig:observador}
\end{figure}

Regarding the fault-tolerant software design, it is composed by two sub-modules: the fault detection and identification module, and the fault-tolerant control module. The former is responsible for fast and accurately detecting and identifying the failure, for which a bank of observers based on \cite{Vey2016} is implemented, and which architecture is shown in Fig. \ref{fig:diag_FDI_v2}. The solution relies on one observer per possible state of the vehicle, one for nominal (detection), and one for each possible rotor in failure (identification). The observer which is consistent with the measured behaviour of the system (that whose residue tends to zero), will be the one that defines in which state the vehicle is. An example of the values of the residues during flight in which a failure is injected is shown in Fig. \ref{fig:observador}.

Regarding the fault-tolerant control, as described before, the vehicle switches between two predetermined states, from nominal to one of the possible failure states. As the geometry of the vehicle changes, so does the control allocation law, which relates the forces exerted by each motor to the vertical thrust and torques exerted by the vehicle. This can be summarized as $[\textbf{q}^T f_z]^T = A \textbf{f}$, where $\textbf{q}\in\mathbb{R}^3$ are the torques, $f_z\in\mathbb{R}_{\geq0}$ the vertical force, $\textbf{f}\in\mathbb{R}^6$ the set of forces exerted by the six motors, and $A\in\mathbb{R}^{4\times 6}$ is the so called torque-force matrix that relates these quantities, depending only on the mechanical characteristics of the vehicle. For example, the force-torque matrix $A$ for the vehicle in nominal state (no failure) in Fig. \ref{fig:hexarotor_platform} is:

\begin{align*}
A &=
\begin{bmatrix}
- \frac{d}{2} & \frac{d}{2}  & d & \frac{d}{2} & - \frac{d}{2} &  - d\\[6pt]
\frac{\sqrt{3}d}{2}  & \frac{\sqrt{3}d}{2} & 0 & - \frac{\sqrt{3}d}{2} &  - \frac{\sqrt{3}d}{2} & 0\\[6pt]
k_t & -k_t & k_t & -k_t & k_t & -k_t \\
-1 & -1 & -1 & -1 & -1 & -1\\
\end{bmatrix}
\end{align*}

\noindent where $d>0$ is the distance of each motor to the geometric center, and $k_t>0$ is a constant of the motors.

Then, when a failure occurs, only the matrix $A$ needs to be changed to consider the lack of the failing motor, and the new orientation of one of the reconfigurable ones. As all these possibilities are pre-computed, the vehicle only switches between a number of predetermined matrices \cite{Pose2020TMech}. This allows to maintain the same control law, such as a PID controller, and keeps the control design simple.

\subsection{Experimental evaluation}

\begin{figure}[t]
    \centering
    \includegraphics[width = \columnwidth]{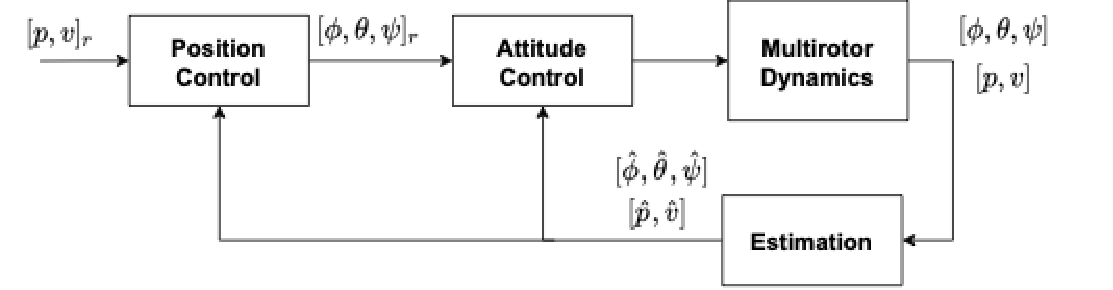}
    \caption{Control scheme diagram attitude and position nested controllers.}
    \label{fig:controllers_diagram}
\end{figure}

The structure of the UAV in Fig. \ref{fig:hexarotor_platform} is based on a DJI F550 Flame Wheel ARF Kit, and the rotors are DJI 2212/920KV Brushless DC motors capable of providing \SI{1}{\kilogram} thrust each, while the total weight of the vehicle sits around \SI{2.8}{\kilogram}. This allows to maintain the vehicle in the air even when one of the rotors is not providing any thrust. 
The UAV is controlled by means of a custom made flight computer \cite{Choriboard}, with a Cortex M3 microcontroller at its core. It includes a variety of sensors, such as an MPU600 Inertial Measurement Unit and a HMC5883L digital compass to estimate the attitude of the vehicle. Two nested PID \cite{astrom2021} controllers are used for position and attitude control, as shown in Fig. \ref{fig:controllers_diagram}.

\begin{figure}[t]
    \centering
    \includegraphics[width = \columnwidth]{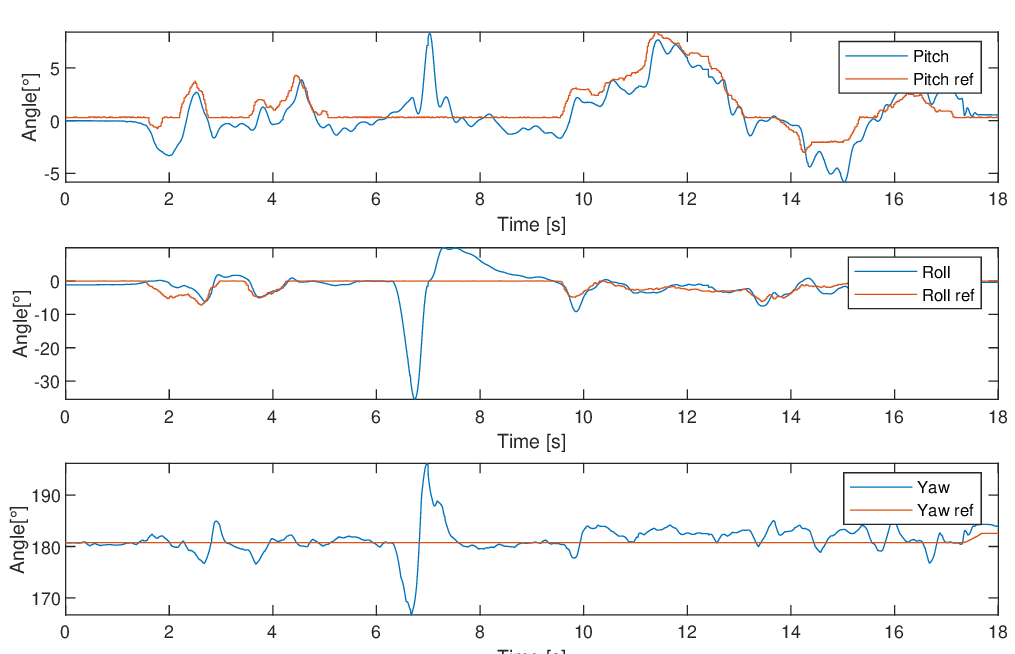}
    \caption{Attitude of the hexarotor in and outdoor flight, where a fault is injected at $t=\SI{6.3}{\second}$.}
    \label{fig:outdoor_fail_atti}
\end{figure}

\begin{figure}[t]
    \centering
    \includegraphics[width = 0.9\columnwidth]{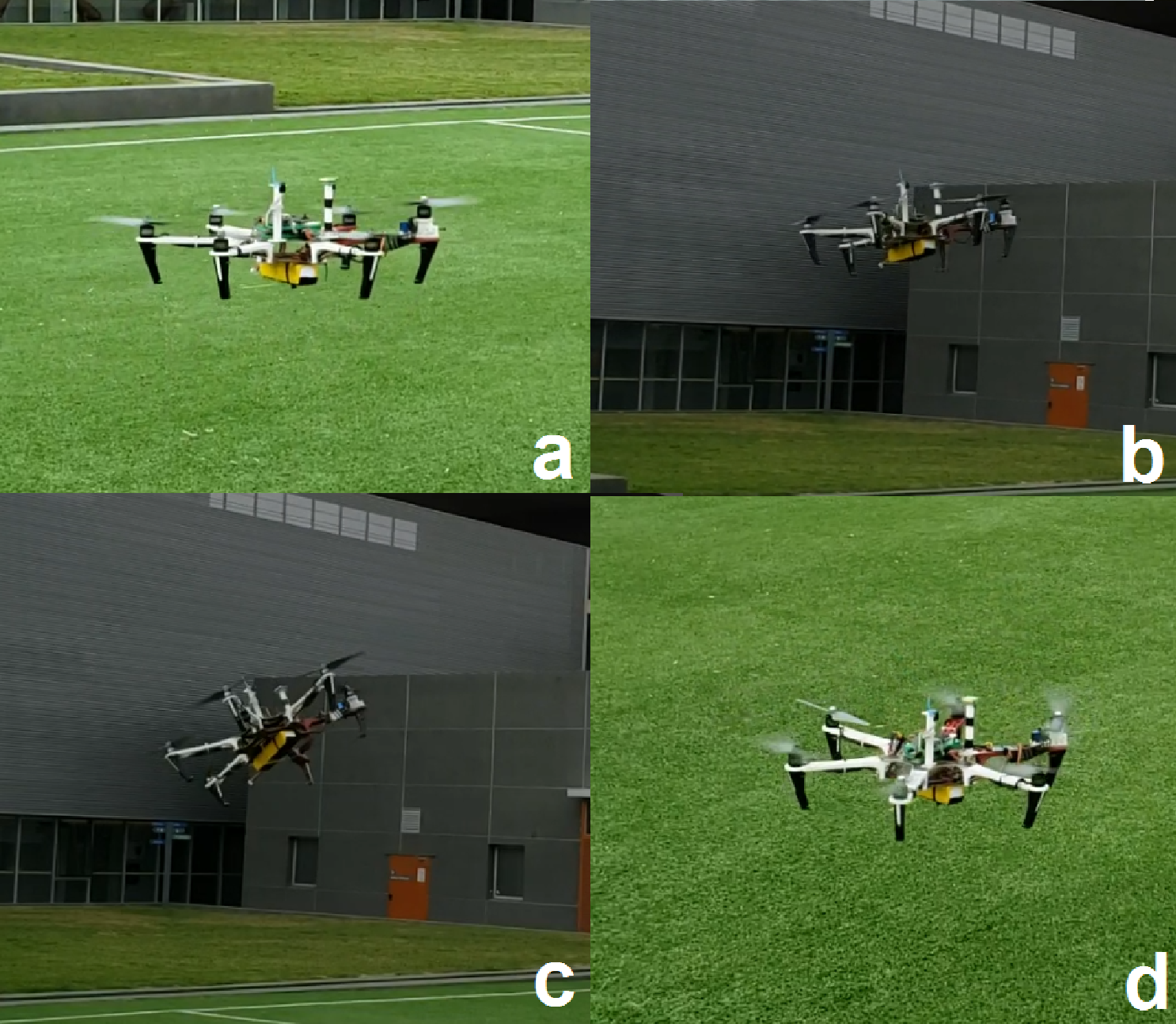}
    \caption{Behavior of the fault-tolerant hexarotor during a failure, detection and recovery maneuver. (a) Vehicle in nominal conditions. (b) One motor stops. (c) The failure is detected and the control allocation is reconfigured. (d) The failure is compensated and the vehicle is stabilized.}
    \label{fig:outdoor_flight}
\end{figure}

In order to provide experimental proof of the fault-tolerant capabilities, several outdoor flights were performed where the vehicle took off without failures, and was injected with a failure in one motor during flight, from which recovers and continues maneuvering without problems.
In Fig. \ref{fig:outdoor_fail_atti} the attitude is shown for the vehicle in and outdoor flight, where a fault is injected at $t=\SI{6.3}{\second}$, detected in around \SI{250}{\milli\second}, and reconfigured, to recover and continue flight till landing. Additionally, Fig. \ref{fig:outdoor_flight} shows the maneuver during the recovery and how it continue flying with one of the motors completely stopped. Some of the experimental flights performed with this vehicle, in outdoor environments, can be seen at \cite{FTCservoOutdoors}.

\section{Camera Pose Estimation}

Estimating the pose of a robot with a camera is a well-studied topic, which is important for any visual-based robotic system. In particular for UAV, attempting to perform a landing maneuver outdoors is quite difficult with standard sensors, e.g. GPS, IMU and magnetometer, not to mention landing in indoor environments where GPS signals cannot be used. Thus other sensors, namely rangefinders, beacons or cameras, are mounted on vehicles to get a better estimate of the vehicle's position. The combination of odometry with visual sensors is capable of achieving a highly-accurate camera pose estimates, see for instance \cite{gautam2014survey} and references therein. 

In order to obtain accurate position estimates from visual data, the modelling of the camera has to be taken into account. One of the most commonly used camera models is the pinhole model, which will be used in this work.

\subsection{Pinhole Model}

\begin{figure}[t]
    \centering
    \includegraphics[width = \columnwidth]{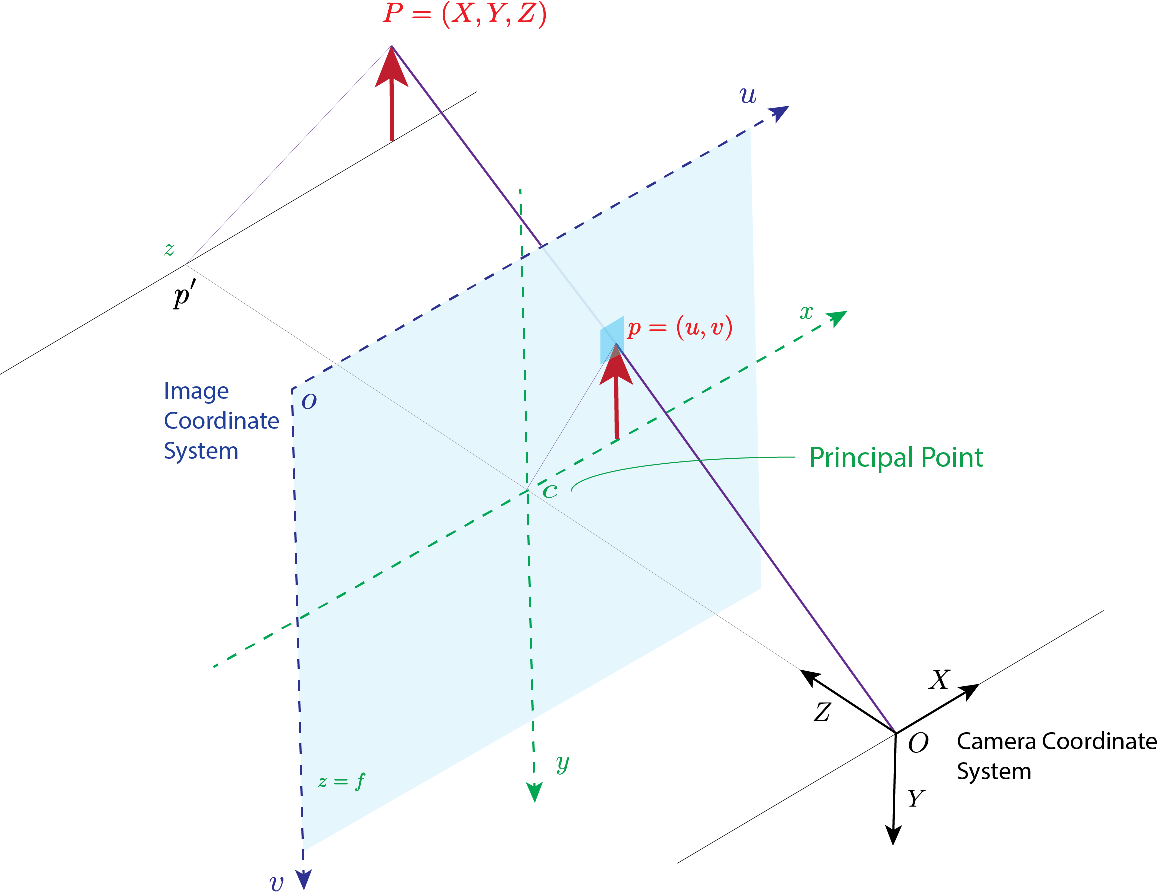}
    \caption{The pinhole model.}
    \label{fig:pinhole_model}
\end{figure}


In Fig. \ref{fig:pinhole_model} the pinhole model is depicted, where the pinhole (or optical center) is defined as a point in space where a single light ray, reflected by a point of the 3D object, passes through and projects said point onto the image plane (retinal plane). The image plane is at a focal distance $f$ from the optical center $O$. 
The camera coordinate system $(X,Y,Z)$ has its optical axis $Z$ pointing towards the image plane, which is at a focal distance $f$ from the optical center. On this plane the image coordinate system $(u,v)$ in blue, and a coordinate system $(x,y)$ at $c$ in green is defined. A light ray coming from a point $\mathbf{P} = (X,Y,Z)$ is projected onto this plane representing a pixel $\mathbf{p} = (u,v)$. Assuming that the projective plane is perpendicular to the Z-axis of the camera coordinate system; the intersection is at the principal point $F = [0,0,f]^T$, expressed in the image coordinate system as $\mathbf{c} = (c_x, c_y)$. 

\subsection{Pose Estimation}


One of the techniques to estimate a vehicle position is the use of fiducial markers. These markers have a predefined shape and size, allowing to estimate depth without using more than one camera. For this work, ArUco tags \cite{Garrido2014} are the markers of choice, which have several convenient properties, such as automatic generation, and self-correction under occlusion. The self-correction property involves detecting the correct marker even if a certain number of bits are corrupted in the image, which can be achieved by occluding a portion of the marker. Deliberately occluding a portion of the marker with a smaller marker will be used as the camera approaches the marker upon landing; thus harnessing this property. 


%
Eq. \eqref{eq:pose_prediction} describes the pose prediction $p_{k+1|k} \in \mathbb{R}^3$ at a given time step $k+1$ based on $k$ previous estimates and velocity $v_k \in \mathbb{R}^3$, where $k \in \mathbb{N}^+$, and $\Delta{t}$ is the time between estimates.

\begin{equation} \label{eq:pose_prediction}
    \mathbf{p}_{k+1|k} = \mathbf{p}_{k|k} + \mathbf{v}_k \Delta{t} 
\end{equation}

Only when there is external position information and the camera pose estimate is achievable, the measured position $\mathbf{y}_k \in \mathbb{R}^3$ is used. Eq. \eqref{eq:pose_update} shows how to update the pose prediction $\mathbf{p}_{k+1|k}$ yielding an update pose estimate $\mathbf{p}_{k+1|k+1}$. The optimal gain $K \in [0,1]$ can be obtained by defining the adequate weights, in order to obtain a good estimation.

\begin{equation} \label{eq:pose_update}
    \mathbf{p}_{k+1|k+1} = \mathbf{p}_{k+1|k} + K (\mathbf{y}_k - \mathbf{p}_{k+1|k})
\end{equation}

In Eq. \eqref{eq:pose_update}, the position of the vehicle w.r.t. the fiducial marker $y_k$ is used, however, fiducial markers algorithms calculate the target position respect to the camera frame. To calculate the vehicle position respect to an inertial frame, a transformation must be applied as follows:

\begin{equation}
    \mathbf{y}_k = -C^i_{b,k} C^b_c \mathbf{p}_c
\end{equation}

\noindent where $C^b_c \in SO(3)$ is the transformation from the camera frame to the vehicle's body frame (fixed), and $C^i_{b,k} \in SO(3)$ the one from the body frame to the inertial frame. 
As the camera cannot provide information about the attitude of the vehicle, it will rely on the vehicle's flight computer in order to obtain this information at each time-step $k$.

In Algorithm \ref{alg:pose_estimation} a pseudocode of the pose estimation based upon visual information is shown. This algorithm is executed at every time-step $k$. Firstly, it must receive information on the periodicity in which the prediction is computed (step 1), the current position of the vehicle, its velocity and attitude given by a rotation matrix. The marker length, and the camera parameters---extracted from the camera calibration---are also needed. Secondly, in step 2, the marker must be detected, to update the position with camera information in step 6-7. Hence, corners are extracted from the image by invoking the marker detection function, which returns an empty array if there is no marker. If the detection is successful, the position of the marker w.r.t. the camera is estimated, and only then the position is updated. 

\begin{algorithm}[t]
\caption{Vision-based pose estimation}\label{alg:pose_estimation}
\begin{algorithmic}[1]
\REQUIRE $\Delta t$, $p$, $v$, $C^b_i$, $C^b_c$, $L$, $K$
\ENSURE $p$
\STATE $p \gets p + v \Delta t$
\STATE corners $\gets$ detectMarkers()
\IF{corners.length() $>$ 0}
    \STATE $t_{tag} \gets$ estimatePoseSingleMarkers(corners, $L$, $K$)
    
    \IF{$t_{tag}$ $\neq \textsc{null}$}
        \STATE $y \gets -C^i_b C^b_c t_{tag}$ 
        \STATE $p \gets p - K(y - p)$
    \ENDIF
\ENDIF

\end{algorithmic}
\end{algorithm}

\subsection{Autonomous landing algorithm validation}

In order to provide proof of concept, several experimental flights were performed in a quadrotor with two on-board computers. One responsible for the low-level control of the vehicle, and the other responsible for the vision subsystem. The former, provides attitude data to the image processing computer, and receives velocity commands from the latter. Both computers will be detailed in-depth in the following sections, for the case of the presented prototype.

\begin{figure}[t]
    \centering
    \includegraphics[width = \columnwidth]{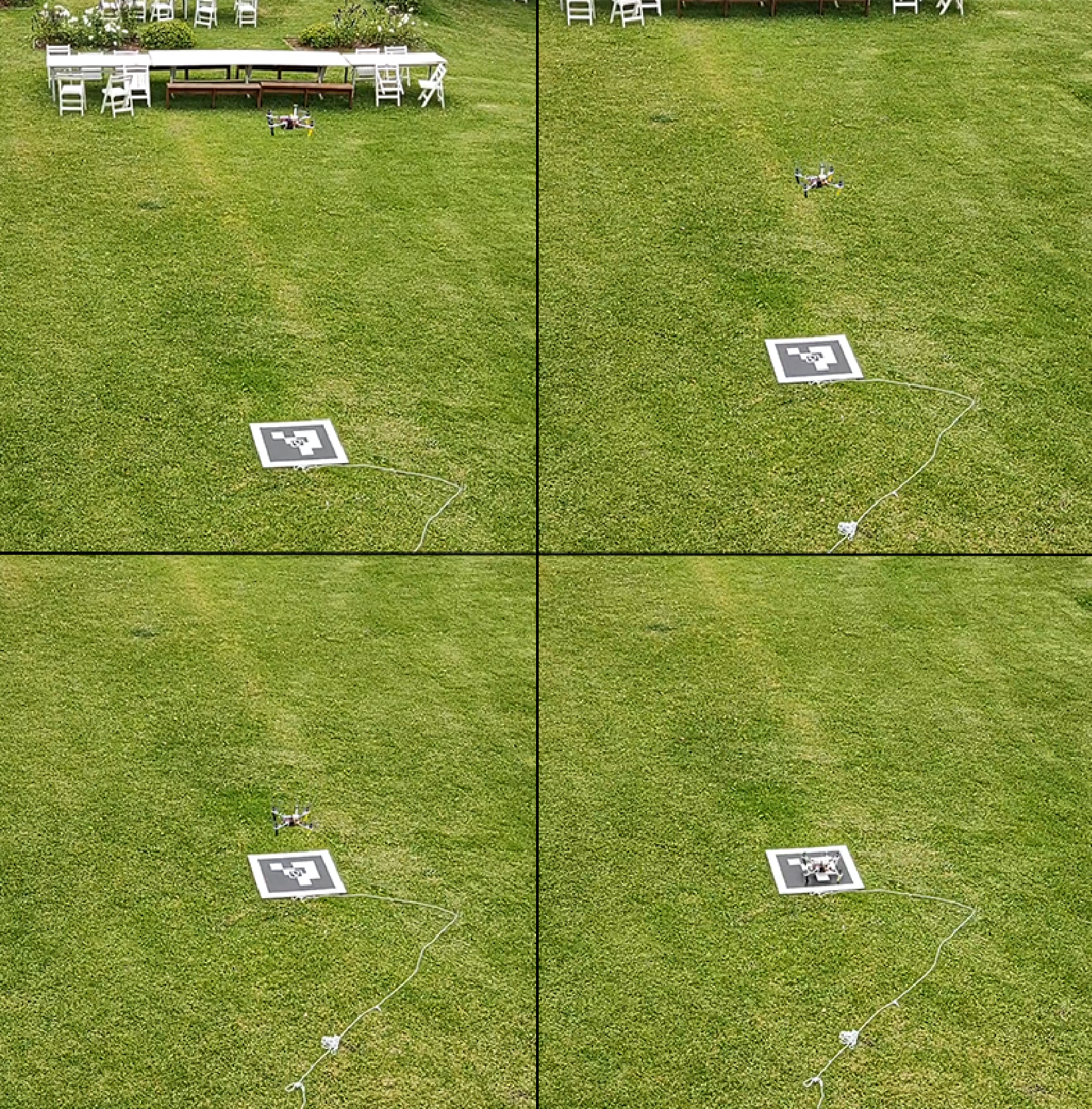}
    \caption{Landing sequence of a multirotor aerial vehicle on top of fiducial marker.}
    \label{fig:land_sequence}
\end{figure}

In Fig. \ref{fig:land_sequence}, the quadcopter is shown performing a landing maneuver, and successfully landing upon a fiducial marker.
The landing sequence is performed after completing a mission consisting in reaching a number of pre-defined waypoints. During the mission, the vehicle follows the route given by those waypoints, the last of which is nearby the desired landing point marked by the fiducial. After reaching it, the vehicle proceeds to execute the landing maneuver. At this time, the image processing computer provides updates on the pose using the information retrieved from the downward camera. In this experiment the landing started at 6 meters above the ground, after making sure the target is in sight at all times. The error w.r.t. the center of the marker at the end of the landing, was around \SI{10}{\centi\meter}-\SI{15}{\centi\meter} under nominal weather conditions. However, the pose estimate using camera information has lower error, and the deviations are mainly caused by the precision in the UAV control. The full flight that ends in the landing maneuver can be seen at \cite{Youtube-quad-landing}.

%

\section{Experimental Prototype}


For this work, an NVIDIA Jetson TX2 Module with a 256-core architecture and 255 NVIDIA CUDA cores is used for runing the image processing algorithms. It is also composed of a Dual-Core NVIDIA Denver 2 64-Bit CPU and a Quad-Core ARM Cortex-A57 MPCore. It is an 8GB LPDDR4 RAM computer with a 1866 MHz operational frequency. In this on-board computer, intensive processing will take place, i.e. image processing and high-level control algorithms. The motivation for using such a powerful computer is due to the flexibility it provides, which will allow to implement more advanced and resource-hungry algorithms in the future, not only image processing algorithms but also fault detection algorithms. An Orbitty Carrier Board is used to provide additional features to the computer, e.g. Wi-Fi, serial connectivity, etc.

The camera of choice is an Intel RealSense Depth Camera SR305. Even though the camera has the capability of providing a channel of depth, it is not used as the application does not rely on such feature, but is one of the possible future implementations. The camera is a high-speed USB camera that can achieve a 60FPS@720p stream of RGB data. It has been experimentally tested that the camera can perceive a fiducial marker up to a distance of 28 meters, which is enough for this work.

As stated before, a custom made computer is tasked with the attitude and position control of the fault-tolerant vehicle. This flight computer is required to send attitude and linear speed to the Jetson TX2 for pose estimation, and need to receive velocity commands in order to reach the landing target. To this end, a bidirectional UART communication channel is implemented between the flight computer and the Jetson TX2 with a custom protocol in order to make the messages as short as possible. The flight computer sends the attitude and linear speed for position estimation as described before, and also position and other variables of interest for data logging purposes. In turn, the Jetson TX2 captures and processes images to obtain a position estimate of the fiducial marker, and generates a velocity command that is sent to the flight computer, allowing to precisely position the vehicle over the landing marker. For high accuracy position estimation when no image data is available, it relies on a GNSS RTK based on the Ardusimple RTK2B.


\begin{figure}[t]
    \centering
    \includegraphics[width = \columnwidth]{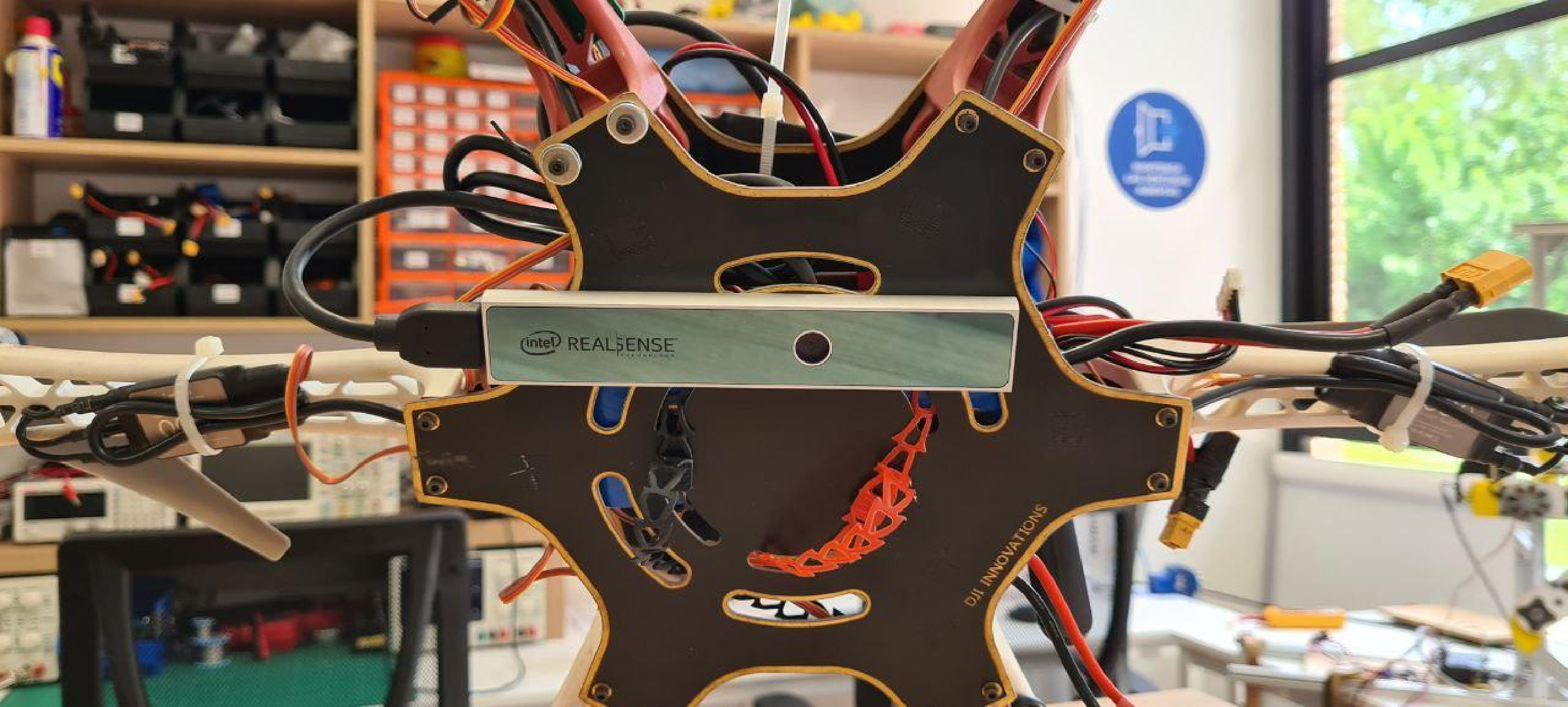}
    \caption{Mounted Intel RealSense Camera on the bottom side of the vehicle.}
    \label{fig:realsense_camera}
\end{figure}

\begin{figure}[t]
    \centering
    \includegraphics[width = \columnwidth]{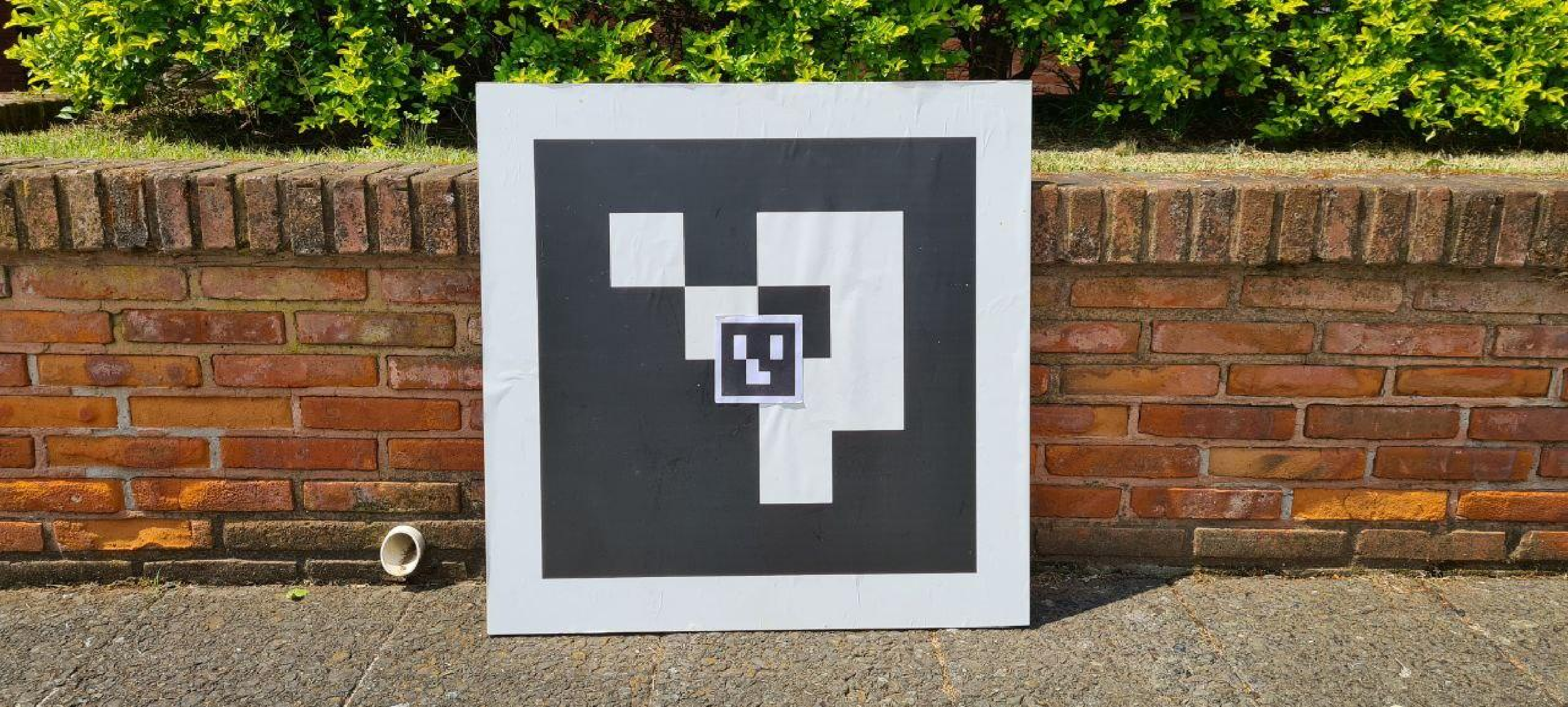}
    \caption{Construction of the landing target based upon two fiducial markers from a 4$\times$4 dictionary. dictionary.}
    \label{fig:landing_target}
\end{figure}

The landing target is the second part of the experimental platform that must be addressed. It consists of a large ArUco fiducial marker and a smaller one. This is to prevent the camera from losing the detection as the vehicle moves toward the ground. Fig. \ref{fig:landing_target} shows how the smaller marker is located in the center of the bigger one. This is possible due to the self-correction algorithm used in the ArUco implementation, making possible to corrupt at most a determined number of bits according to the properties of the marker. The big and small markers are 640 mm and 108 mm in length, respectively; the markers are placed on top of a 800 mm by 800 mm MDF board to avoid moving the landing platform as air flows are generated by the UAV, upon either take-off or landing. 

The NVIDIA Jetson TX2 Module will run on Ubuntu 20.04 and ROS Noetic, as these distributions are the most stable ones. The landing controller will be implemented in Python and in a multi-threaded context. Even though a single node is used to accomplish our goal, it is quite useful for various reasons, i.e. node monitoring and multi-threading capabilities.

Multi-threading is desired to allow the image acquisition thread to acquire images as fast as it can, without worrying about other threads, such as the control loop, marker detection, and pose estimation. The image acquisition thread will run as fast as the camera can acquire images at a given resolution.



\balance
\section{Discussion \& Future Work}



This work has presented a prototype for an autonomous precision landing hexarotor with fault-tolerant capabilities. Both parts, as separate entities, have achieved a good performance. On one hand, the fault-tolerant vehicle has shown very good maneuverability after a failure has occurred, even considering the additional weight due to the addition of the Jetson TX2 and the camera. On the other hand, the image processing algorithm has shown an efficient implementation in the Jetson TX2 with good frame rate, and could achieve a precise landing when implemented in a quadrotor controlled by a Pixhawk.

As the time of writing this paper, there have been some experimental tests involving the integrated system. However, these tests are not conclusive yet. In the future it is planed to carry out
experimental test performing the landing maneuver in nominal conditions, and then repeat these tests in a total failure condition. After the vehicle performs satisfactorily in both cases, a final test will be performed in where the failure is injected during the landing maneuver. This will cause the vehicle to abruptly move, and will allow to test in depth the prototype, as situations such as loss of target may happen.

A planning algorithm will be implemented, capable of aborting the landing maneuver when the failure occurs if necessary. When the vehicle is recovered form the failure, will restart the landing maneuver. 
Several questions remains, for instance what happens if the failure occurs near the floor, if it is possible to define an altitude to guarantee that the vehicle will be capable of recover from the failure.

Additionally, the system presented here also works if the target is moving, assuming that is in the vehicle field of view. But, if the target moves out of the camera field of view, the UAV will fail to land on top of it. For this, estimating the state of the moving target and predicting its position in future time-steps would solve the moving target problem, several algorithms are proposed in the literature, and some of them will be tested in order to evaluate its performance for certain applications. 

\bibliographystyle{IEEEtran}
\bibliography{bibGPSIC.bib}

\begin{thebibliography}{10}
\providecommand{\url}[1]{#1}
\csname url@samestyle\endcsname
\providecommand{\newblock}{\relax}
\providecommand{\bibinfo}[2]{#2}
\providecommand{\BIBentrySTDinterwordspacing}{\spaceskip=0pt\relax}
\providecommand{\BIBentryALTinterwordstretchfactor}{4}
\providecommand{\BIBentryALTinterwordspacing}{\spaceskip=\fontdimen2\font plus
\BIBentryALTinterwordstretchfactor\fontdimen3\font minus \fontdimen4\font\relax}
\providecommand{\BIBforeignlanguage}[2]{{%
\expandafter\ifx\csname l@#1\endcsname\relax
\typeout{** WARNING: IEEEtran.bst: No hyphenation pattern has been}%
\typeout{** loaded for the language `#1'. Using the pattern for}%
\typeout{** the default language instead.}%
\else
\language=\csname l@#1\endcsname
\fi
#2}}
\providecommand{\BIBdecl}{\relax}
\BIBdecl

\bibitem{AAM1}
\BIBentryALTinterwordspacing
R.~Goyal, C.~Reiche, C.~Fernando, and A.~Cohen, ``Advanced air mobility: Demand analysis and market potential of the airport shuttle and air taxi markets,'' \emph{Sustainability}, vol.~13, no.~13, 2021. [Online]. Available: \url{https://www.mdpi.com/2071-1050/13/13/7421}
\BIBentrySTDinterwordspacing

\bibitem{AAMFund1}
C.~Dietrich, T.~Johnston, and R.~Riedel, ``{Looking to the skies: Funding for future air mobility takes off},'' \emph{McKinsey {\&} Company}, 2021.

\bibitem{NASAAAM}
``{Regional Air Mobility: Leveraging our National Investments to Energize the American Travel Experience},'' NASA Langley Research Center, Tech. Rep., 2021.

\bibitem{giribet2016analysis}
J.~I. Giribet, R.~S. Sanchez-Pe\~na, and A.~S. Ghersin, ``Analysis and design of a tilted rotor hexacopter for fault tolerance,'' \emph{IEEE Transactions on Aerospace and Electronic Systems}, vol.~52, no.~4, pp. 1555--1567, 2016.

\bibitem{giribet2018experimental}
J.~I. Giribet, C.~D. Pose, A.~S. Ghersin, and I.~Mas, ``Experimental validation of a fault tolerant hexacopter with tilted rotors,'' \emph{International Journal of Electrical and Electronic Engineering and Telecommunications}, vol.~7, no.~2, pp. 1203--1218, 2018.

\bibitem{Michieletto2017}
G.~{Michieletto}, M.~{Ryll}, and A.~{Franchi}, ``Control of statically hoverable multi-rotor aerial vehicles and application to rotor-failure robustness for hexarotors,'' in \emph{2017 IEEE International Conference on Robotics and Automation (ICRA)}, May 2017, pp. 2747--2752.

\bibitem{Pose2020TMech}
C.~D. Pose, J.~I. Giribet, and I.~Mas, ``Fault tolerance analysis for a class of reconfigurable aerial hexarotor vehicles,'' \emph{IEEE/ASME Transactions on Mechatronics}, vol.~25, no.~4, pp. 1851--1858, 2020.

\bibitem{Vey2016}
D.~Vey and J.~Lunze, ``Experimental evaluation of an active fault-tolerant control scheme for multirotor {UAVs},'' \emph{3rd International Conference on Control and Fault-Tolerant Systems}, pp. 119--126, 2016.

\bibitem{Choriboard}
L.~Garberoglio, M.~Meraviglia, C.~D. Pose, J.~I. Giribet, and I.~Mas, ``Choriboard iii: A small and powerful flight controller for autonomous vehicles,'' in \emph{2018 Argentine Conference on Automatic Control (AADECA)}, 2018, pp. 1--6.

\bibitem{astrom2021}
A.~K. J. and R.~M. Murray, \emph{Feedback systems: An introduction for scientists and Engineers}.\hskip 1em plus 0.5em minus 0.4em\relax Princeton University Press, 2021.

\bibitem{FTCservoOutdoors}
\BIBentryALTinterwordspacing
``{Fault tolerant hexacopter - Outdoor flight test}.'' [Online]. Available: \url{https://www.youtube.com/watch?v=9CGn-OY6JQk}
\BIBentrySTDinterwordspacing

\bibitem{gautam2014survey}
A.~Gautam, P.~Sujit, and S.~Saripalli, ``A survey of autonomous landing techniques for uavs,'' in \emph{2014 international conference on unmanned aircraft systems (ICUAS)}.\hskip 1em plus 0.5em minus 0.4em\relax IEEE, 2014, pp. 1210--1218.

\bibitem{Garrido2014}
S.~Garrido-Jurado, R.~Muñoz-Salinas, F.~Madrid-Cuevas, and M.~Marín-Jiménez, ``Automatic generation and detection of highly reliable fiducial markers under occlusion,'' \emph{Pattern Recognition}, vol.~47, p. 2280–2292, 06 2014.

\bibitem{Youtube-quad-landing}
\BIBentryALTinterwordspacing
``{Quadrotor vision-based precision landing}.'' [Online]. Available: \url{https://www.youtube.com/watch?v=DSj5nErO3x8}
\BIBentrySTDinterwordspacing

\end{thebibliography}
\end{document}